\documentclass[letterpaper]{article} 
\usepackage{aaai23}  
\usepackage{times}  
\usepackage{helvet}  
\usepackage{courier}  
\usepackage[hyphens]{url}  
\usepackage{graphicx} 
\usepackage{natbib}  
\usepackage{caption} 
\usepackage{hyperref}
\usepackage{xr}
\usepackage{threeparttable}
\usepackage{bbding}
\usepackage{bm}
\usepackage{amssymb}
\usepackage{amsmath}
\newcommand{\eg}{\textit{e}.\textit{g}.}
\newcommand{\ie}{\textit{i}.\textit{e}.}
\newcommand{\etal}{\emph{et al}.}
\usepackage{subcaption}
\usepackage{adjustbox}
\usepackage{booktabs}
\usepackage{multirow}

\title{Open-Vocabulary Multi-Label Classification via Multi-Modal Knowledge Transfer}
\author{
    Sunan He\textsuperscript{\rm 1,2,3}\equalcontrib\thanks{This work was done during an internship at Tencent.},
    Taian Guo\textsuperscript{\rm 3}\equalcontrib, 
    Tao Dai\textsuperscript{\rm 1}\thanks{Corresponding author: Tao Dai},
    Ruizhi Qiao\textsuperscript{\rm 3},
    Xiujun Shu\textsuperscript{\rm 3},
    Bo Ren\textsuperscript{\rm 3},
    Shu-Tao Xia\textsuperscript{\rm 2,4}
}
\affiliations{
      \textsuperscript{\rm 1} College of Computer Science and Software Engineering, Shenzhen University\\
    \textsuperscript{\rm 2} Tsinghua Shenzhen International Graduate School, Tsinghua University \\
    \textsuperscript{\rm 3} YouTu Lab, Tencent 
    \textsuperscript{\rm 4} Research Center of Artificial Intelligence, Peng Cheng Laboratory\\
    hsn20@mails.tsinghua.edu.cn, \{taianguo, ruizhiqiao, xiujunshu, timren\}@tencent.com \\
    daitao.edu@gmail.com, xiast@sz.tsinghua.edu.cn
}

\begin{document}

\maketitle

\begin{abstract}
Real-world recognition system often encounters the challenge of unseen labels.
To identify such unseen labels, multi-label zero-shot learning (ML-ZSL) focuses on transferring knowledge by a pre-trained textual label embedding (\eg, GloVe).
However, such methods only exploit \textit{single-modal} knowledge from a language model, while ignoring the rich semantic information inherent in image-text pairs.
Instead, recently developed open-vocabulary (OV) based methods  succeed in exploiting such information of image-text pairs  in  object detection, and achieve impressive performance.
Inspired by the success of OV-based methods, we propose a novel open-vocabulary framework, named multi-modal knowledge transfer (MKT), for  multi-label classification.
Specifically, our method exploits \textit{multi-modal} knowledge of image-text pairs  based on a vision and language pre-training (VLP) model.
To facilitate transferring the image-text matching ability of VLP model, knowledge distillation is employed to guarantee the consistency of image and label embeddings, along with prompt tuning to further update the label embeddings.
To further enable the recognition of multiple objects, a simple but effective two-stream module is developed to capture both local and global features.
Extensive experimental results show that our method significantly outperforms state-of-the-art methods on public benchmark datasets. 
The source code is available at \href{https://github.com/sunanhe/MKT}{https://github.com/sunanhe/MKT}.
\end{abstract}

\section{Introduction}
Multi-label recognition, which aims to recognize all the relevant labels in an image, is a fundamental task in computer vision applications, such as scene understanding, surveillance systems and self-driving cars.
In real-world applications, multi-label recognition systems should learn tens of thousands of labels, locate them in images, and even deal with many unseen labels.
To date, classic multi-label classification methods trained and tested with seen labels are far from fulfilling the requirements for real applications, where plenty of unseen labels exist.
\par
\begin{figure}[t]
\centering
\includegraphics[width=1\linewidth, page=6]{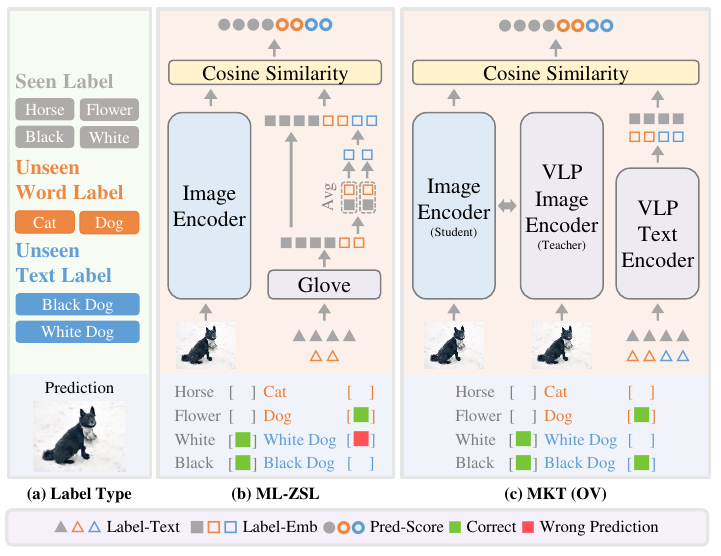}
\caption{The overall framework of the classic multi-label zero-shot learning (ML-ZSL), and our multi-modal knowledge transfer (MTK) method. (b) ML-ZSL only exploits single-modal knowledge of language-based models (\eg, Glove), and may fail to recognize unseen text labels (\eg, `Black Dog'). (c) Instead, our MKT  succeeds in predicting it by jointly exploring multi-modal knowledge of vision and language pre-training (VLP) models. (Best viewed in color.)
}
\label{fig:intro}
\end{figure}

To identify the unseen labels in an image,
many multi-label zero-shot learning (ML-ZSL) methods~\cite{lesa, ganmlzsl, zssdl, biam} have been recently developed by transferring  knowledge between  seen and unseen labels.
However, most existing  methods~\cite{fast0tag,lesa,ganmlzsl,zssdl,biam} 
contain two main issues. 
\textbf{First},  these methods solely exploit \textit{single-modal} knowledge by a pre-trained textual label embeddings like GloVe~\cite{glove} (as shown in Figure~\ref{fig:intro} (b)),
while ignoring the visual semantic image-text pair information.
 \textbf{Second}, although such textual label embeddings  (\eg, GloVe) handle  word labels (\eg, label of `cat') well,  they cannot be easily  extended to text labels (\eg, label of `black cat'), thus hindering the flexibility of the models. 
As shown in Figure~\ref{fig:intro}, ML-ZSL fails to recognize the unseen text label of `black dog', while our MKT succeeds in predicting this label by jointly exploring \textit{multi-modal} knowledge of vision and language models.

To explore such multi-modal knowledge, recently developed open-vocabulary (OV) methods \cite{openvild, opensedo, openseg, openprompt, liang2022mind} have been proposed  based on vision and language pre-training (VLP) models.
Such OV-based methods trained on billions of image-text pairs contain powerful image-text matching ability, and have achieved remarkable performance in computer vision tasks like object detection.
However,  how to extend such OV-based methods  to multi-label classification, including unseen text labels, is less explored.
\par
Motivated by the above observations, 
we propose a novel open-vocabulary framework, named multi-modal knowledge transfer (MKT), for multi-label classification.
Unlike the previous ML-ZSL methods that exploit only language-based information, our MKT utilizes \textit{multi-modal} knowledge from image-text pairs from a  vision and language pre-training (VLP) model.
As shown in Figure~\ref{fig:intro}(c), our MKT mainly consists of an image encoder to extract image features,  and a VLP image/text encoder
to extract image/label embeddings.
Specifically,  
to facilitate transferring the image-text matching ability of VLP models, 
knowledge distillation and prompt tuning are introduced to guarantee the consistency of image and label embeddings.
In practice, knowledge distillation  makes image embeddings align better with its relevant label embeddings, while prompt tuning adapts the label embeddings to better support classification task.
Besides, to further improve the ability of feature  expressions,
we propose a simple but effective two-stream feature extraction module to capture both local and global features to extract more discriminative features.
In this way, our MKT framework can capture the rich semantic information inherent in image-text pairs of VLP models.
\par
The main contributions can be summarized as follows:
\begin{enumerate}
    \item We propose an open-vocabulary based multi-modal knowledge transfer (MKT) framework for multi-label classification, which exploits the semantic multi-modal information in image-text pairs based on VLP models.
    To the best of our knowledge, this is the first work to explore open-vocabulary multi-label classification task.
    \item Our MKT framework mainly consists of an image encoder to extract image features, and a VLP image/text encoder to extract image/label embeddings.
    To guarantee the consistency of image and label embeddings, a knowledge distillation strategy is incorporated into our MKT framework, along with prompt tuning to update the label embeddings iteratively.   Besides, to further improve the ability of feature expressions of our method, 
    we propose a two-stream feature extraction module by  jointly capturing local and global features. 
    \item Extensive results show that our MKT method significantly outperforms the previous ML-ZSL methods and establishes a new state of the art for open-vocabulary multi-label classification on two large-scale benchmark datasets, namely NUS-WIDE and Open Images.
\end{enumerate}
\begin{figure*}[ht]
    \centering
    \includegraphics[width=0.9\textwidth, page=2]{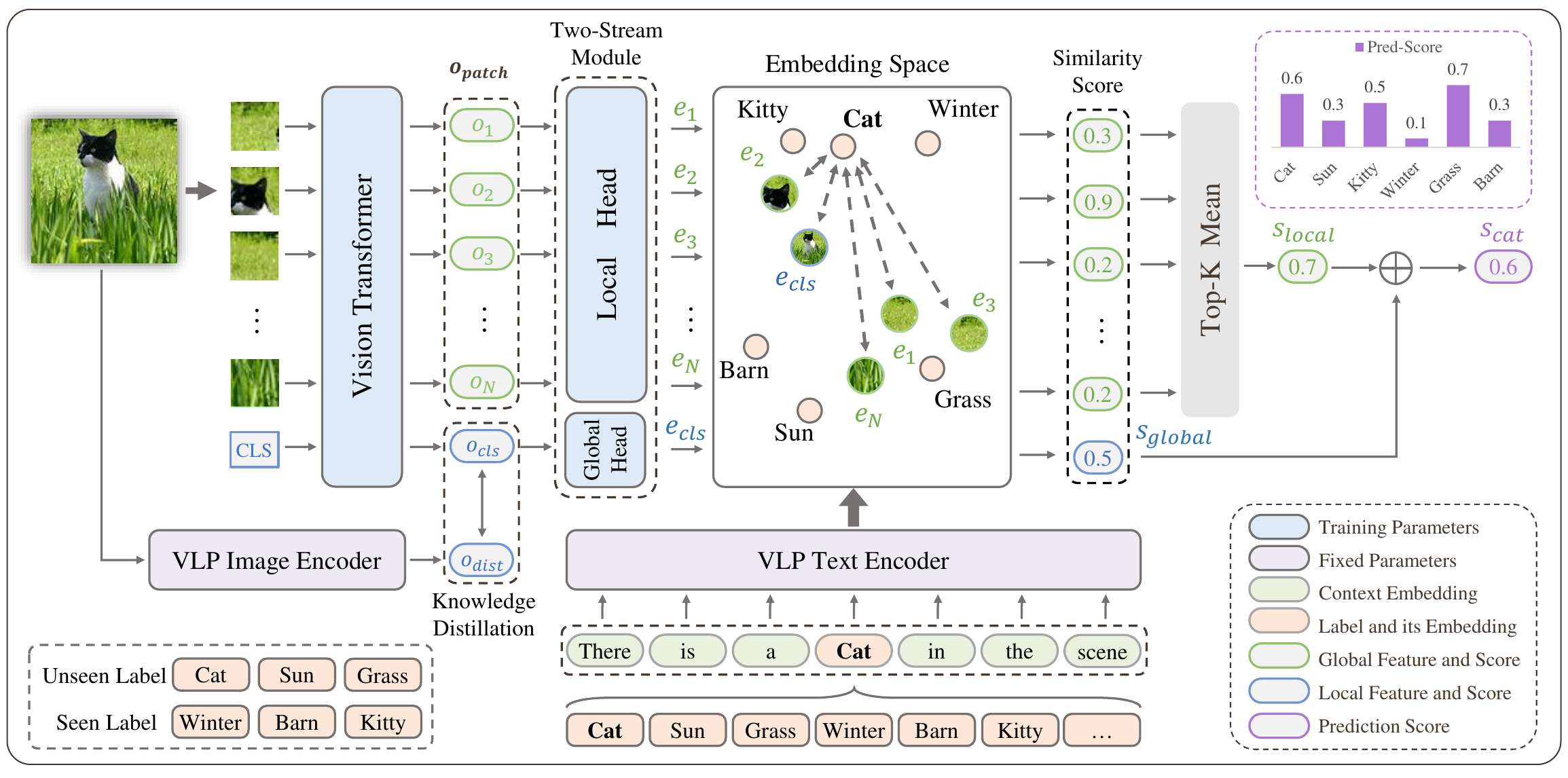}
    \caption{The overall framework of our multi-modal knowledge transfer (MKT) model for open-vocabulary multi-label classification.
    Our MKT mainly consists of a  vision and language pre-training (VLP) model and a vision transformer model.
    The VLP model aims to extract multi-modal knowledge of  input image-text pairs, while vision transformer is used to extract semantic features of input images. 
    Moreover, knowledge distillation is used to guarantee the consistency of image and its relevant label embeddings, along with prompt tuning to further update the label embeddings.
    (Best viewed in color.)
    }
    \label{fig:overall_arch}
\end{figure*}
\section{Related Works}

\subsection{Multi-Label Zero-Shot Learning}
The goal of standard multi-label classification task is to predict a set of labels in an image. 
A vanilla approach is to train a binary classifier for each label present in the training dataset without considering the dependence among the labels~\cite{tsoumakas2007multi, read2011classifier}. 
To capture the label correlation, structure learning~\cite{gong2014deep,Wang_2016_CVPR,Zhu_2017_CVPR,Wang_2017_ICCV} and graph methods~\cite{Li_2016_CVPR,Lee_2018_CVPR,Chen_2019_CVPR} are introduced in this task. 
Recently, vision transformer based methods have received much attention due to the powerful ability of  capturing the global dependency~\cite{lanchantin2021general,liu2021query2label,cheng2021mltr}.
Although these methods have achieved promising results in multi-label classification, they cannot handle unseen labels, thus limiting their real applications. \par
To identify the unseen labels, zero-shot learning (ZSL) usually utilizes semantic information like attributes or word embeddings\cite{mikolov2013distributed,Xian_2017_CVPR}.
In particular,
Lampert \etal~\cite{5206594} proposed two attribute-based paradigms with direct attribute prediction (DAP) and indirect attribute prediction (IAP). 
The former aims to learn multiple attribute classifiers~\cite{6571196}, while the latter uses seen class proportions for prediction~\cite{Zhang_2015_ICCV}. 
While they can recognize to single unseen label, they cannot handle multi-label problem. 
\par
As an extension of ZSL,
multi-label zero-shot learning (ML-ZSL) is developed to identify multiple seen and unseen labels in an image.
The keys to this task are the alignment of image embeddings with its relevant label embeddings and the relation between seen and unseen label embeddings.
To this end, Fast0Tag~\cite{fast0tag} and ZS-SDL~\cite{zssdl} aim to find principal directions of an image along which the relevant labels rank higher.
LESA~\cite{lesa} and BiAM~\cite{biam} introduce attention module to capture both local and global features for better recognition of multiple objects.
On the other hand, GAN-MLZSL~\cite{ganmlzsl} introduces generative adversarial networks (GANs) to tackle the problem of multi-label feature synthesis from corresponding multi-label class embedding. \par
However, most existing ML-ZSL works exploit only single-modal knowledge via a language model(\eg, GloVe).
Due to the lack of visual information, these language-based models cannot capture visual consistency among labels, thus limiting the generalization ability.
By contrast, we attempt to explore multi-modal knowledge from VLP models to leverage the consistency of image and label embeddings and  can handle multiple word and text unseen labels.

\subsection{Open-Vocabulary Classification} 
With recent great development in vision and language pre-training model, open-vocabulary  classification emerges as an alternative way to predict arbitrary labels.
Large-scale pre-trained models first become prevalent in natural language processing (NLP), such as BERT~\cite{devlin2018bert} and GPT2~\cite{radford2019language}.
Based on large-scale language corpus~\cite{raffel2020exploring} and multiple task-agnostic pre-training objectives~\cite{devlin2018bert}, these pre-trained models achieve promising results in downstream tasks.
Recently, Vision and Language Pre-training (VLP) models~\cite{ViLBERT, chen2020uniter, li2020unicoder, li2020oscar, kim2021vilt} have received much attention in multi-modal tasks. 
For example, with billions of image-text pairs as training samples, CLIP~\cite{clip} and ALIGN~\cite{align} have achieved impressive performance in image-text matching task.
By transferring this matching ability to the classification task, we can achieve arbitrary text label prediction.
Specifically, for any concept, we can generate its label embedding through the text encoder of VLP model and calculate its similarity to image embedding for classification.
Due to the large scale training corpus, we can excavate label embedding of an unbounded vocabulary and achieve open-vocabulary (OV) classification. \par
Some works have explored the OV classification in object detection~\cite{ovrcnn, openvild, openprompt, openonstage, opendetr} and image segmentation~\cite{opensedo, openseg}. 
They usually replace the classification head with label embeddings and achieve impressive performance in arbitrary text concept recognition. 
Moreover, to boost the classification ability, knowledge distillation~\cite{hinton2015distilling} and prompt tuning~\cite{li2021prefix} are introduced to facilitate transferring the image-text matching ability~\cite{coop}. \par
However, most existing OV works focus on single label classification task.
Multi-label classification is more practical and challenging because the models need to recognize multiple objects and cannot be trained with contrastive loss directly.
In this work, we first explore the multi-label open-vocabulary classification task and propose a novel multi-modal knowledge transfer (MKT) framework by jointly exploiting multi-modal knowledge of the image-text pairs based on vision and language pre-training models.  

\section{Multi-modal Knowledge Transfer}

\subsection{Preliminary}
Similar to the ML-ZSL problem, suppose we have two disjoint label sets $Y^{S}$ and $Y^{U}$, where $Y^{S}$ denotes seen labels present in the training set and $Y^{U}$ denotes unseen labels without training images. 
Let $\left(\mathbf{x}_{1}, \mathbf{y}_{1}\right), \ldots,\left(\mathbf{x}_{N}, \mathbf{y}_{N}\right)$ be $N$ training sample, where $\mathbf{x}_{i}$ denotes the $i$-th training samples and $\mathbf{y}_{i} \in Y^{S}$ denotes the labels present in the image.
In the standard zero-shot learning (ZSL) task, the goal is to learn a classifier $f_{Z S L}: X \rightarrow Y^{U}$ to identify the relevant unseen labels for a given image. 
Note that in a more challenging and realistic setup of generalized zero-shot learning (GZSL) task, the classifier needs to identify both seen and unseen labels present in the test image, i.e., $f_{G Z S L}: X \rightarrow Y^{U} \cup Y^{S}$. \par

\subsection{The Overall Framework}
As illustrated in Figure~\ref{fig:overall_arch}, we show the overall architecture of our multi-modal knowledge transfer (MKT) method, which mainly consists of a vision transformer and a vision and language pre-training (VLP) model. 
Specifically,
We utilize the vision transformer~\cite{vit} as our  backbone network to extract semantic features from input images.
Due to its powerful ability in learning visual representations,
we choose CLIP~\cite{clip} as our VLP model to extract semantic multi-modal knowledge from both VLP image and text encoders.
Concretely, the label embedding is first generated based on the VLP text encoder, followed by further updates through prompt tuning.
Moreover, knowledge distillation is introduced to facilitate the alignment between image embeddings and its relevant labels.
\subsection{Vision Transformer with Two-Stream Module}
Denote an input image as $\mathbf{x} \in \mathbb{R}^{C \times H \times W}$, where $H\times W$ is the size of the image and $C$ is the number of channels. 
Following~\cite{vit}, we reshape it into a sequence of flattened 2D patches $\mathbf{x}_{patch} \in \mathbb{R}^{N \times\left(P^{2} \cdot C\right)}$, where $P$ denotes the size of each patch and the total number of patches is $N=H W / P^{2}$.
Followed by a trainable linear projection, $\mathbf{x}_{patch}$ is mapped into $\Bar{\mathbf{x}}_{patch} \in \mathbb{R}^{N \times D}$, where $D$ is input embedding dimension.
Then the processing of the $k$-th block in vision transformer is formulated as
\begin{equation}
\begin{aligned}
&\mathbf{x}_{0}=\left[\mathbf{E}_{cls}, \Bar{\mathbf{x}}_{\text {patch}}\right]+\mathbf{E}_{\text {pos}}, \\
&\mathbf{y}_{k}=\mathbf{x}_{k-1}+\operatorname{MSA}\left(\operatorname{NORM}\left(\mathbf{x}_{k-1}\right)\right), \\
&\mathbf{x}_{k}=\mathbf{y}_{k}+\operatorname{MLP}\left(\operatorname{NORM}\left(\mathbf{y}_{k}\right)\right),
\end{aligned}
\end{equation}
where $\mathbf{E}_{cls}$ is the class token embedding and $\mathbf{E}_{pos}$ is the position embedding. 
$\left[\cdot, \cdot\right]$ means concatenation.
$\operatorname{MLP}\left(\cdot\right)$, $\operatorname{NORM}\left(\cdot\right)$, $\operatorname{MSA}\left(\cdot\right)$ denote multilayer perceptron, norm layer and multi-head self-attention, respectively.

Denote the output of vision transformer as $\mathbf{x}_{L}= \left[\mathbf{o}_{cls}, \mathbf{o}_{patch}\right]$, where $\mathbf{o}_{cls}$ and $\mathbf{o}_{patch}$ correspond to the output of class and patch tokens, respectively. 
$\mathbf{o}_{cls}$ represents the global feature and $\mathbf{o}_{patch}$ denotes the local features.
\par
To identify multiple labels in an image, we propose a simple two-stream module consisting of local head $\Theta_{L}\left(\cdot\right)$ and global head $\Theta_{G}\left(\cdot\right)$, mapping local and global features into embedding space respectively, \par
\begin{equation}
\begin{aligned}
\mathbf{e}_{cls}=\operatorname{\Theta_{G}}\left(\mathbf{o}_{cls}\right), \mathbf{e}_{patch}=\operatorname{\Theta_{L}}\left(\mathbf{o}_{patch}\right), 
\end{aligned}
\end{equation}
where $\mathbf{e}_{patch}= [\mathbf{e}_{1}, \mathbf{e}_{2}, \dots, \mathbf{e}_{N}]$ and $\mathbf{e}_{cls}$ are local and global feature embeddings respectively. \par
Then, final prediction score is formulated as
\begin{equation}
s_{i}=\frac{1}{2}\left\langle\mathbf{z}_{i}, \mathbf{e}_{cls}\right\rangle + \frac{1}{2}\operatorname{TopK}\left(\left[\left\langle\mathbf{z}_{i}, \mathbf{e}_{1}\right\rangle,
\left\langle\mathbf{z}_{i}, \mathbf{e}_{2}\right\rangle,
...,
\left\langle\mathbf{z}_{i}, \mathbf{e}_{N}\right\rangle
\right]\right), 
\label{eq:logit}
\end{equation}
where $\mathbf{z}_{i} \in \mathbb{R}^{1 \times D_{e}}$ is a label embedding and $\operatorname{TopK} \left(\cdot\right)$ is the \textit{top}-$k$ mean pooling. $\langle\cdot, \cdot\rangle$ denotes inner product. \par
The ranking loss $\mathcal{L}_{\text {rank }}$ on prediction scores are used to train the network:
\begin{equation}
\mathcal{L}_{\text {rank }} \triangleq \sum_{i} \sum_{p \in \mathbf{y}_{i} , n \notin \mathbf{y}_{i}} \max \left(1+s_{i}^{n}-s_{i}^{p}, 0\right),
\end{equation}
where $\mathbf{y}_{i} \in Y^{S}$ is the target labels of an image $i$. $s_{i}^{n}$ and $s_{i}^{p}$ denote the scores of negative and positive labels.

\subsection{Knowledge Distillation for Alignment}
As a key point to generalize to unseen labels, the alignment of an image embedding with its associated seen label embeddings plays a critical role in open-vocabulary classification.
We take CLIP~\cite{clip} as our VLP model, consisting of an image encoder and a text encoder.
Considering that the pre-training task of CLIP is to match the paired image and text, the image embedding generated by the CLIP image encoder should be similar to its relevant label embeddings generated by the CLIP text encoder.
Thus, we introduce knowledge distillation to facilitate the alignment between the embeddings of an image and its relevant labels.

Denote the teacher model (\ie, CLIP image encoder) as $\Phi_{I}^{CLIP}\left(\cdot\right)$, then the process of distillation is formulated as
\begin{equation}
\mathcal{L}_{dist} \triangleq \left\| \Phi_{I}^{CLIP}\left(\mathbf{x}\right) - \mathbf{o}_{cls} \right\|_{1} = \left\| \mathbf{o}_{dist} - \mathbf{o}_{cls} \right\|_{1},
\end{equation}
where $\mathbf{x}$ is an image input, $\mathbf{o}_{cls}$ is the global features generated by the student model (\ie, our vision backbone), and $\mathbf{o}_{dist}$ denotes the output of CLIP image encoder.
The reason for distillation on the global features instead of the local is twofold. 
First, both $\mathbf{o}_{cls}$ and the output of CLIP image encoder are corresponded to the CLS token.
Moreover, the local features $\mathbf{o}_{patch}$ corresponding to different input patches are expected to be discriminative instead of identical in order to facilitate the recognition of multiple objects.   

\begin{table*}[t]
\centering

\adjustbox{width=\linewidth}{
\begin{tabular}{c|c|cccccccc|cccccccc} 
\toprule[0.15em]
\multirow{3}{*}{\textbf{ Method}}& \multirow{3}{*}{\textbf{Setting}} & \multirow{3}{*}{\textbf{Task }} & \multicolumn{7}{c}{\textbf{NUS-WIDE ( \#seen / \#unseen = 925/81) }} & \multicolumn{8}{c}{\textbf{Open-Images ( \#seen / \#unseen = 7186/400) }} \\
 &  &  & \multicolumn{3}{c}{\textbf{K = 3 }} & \multicolumn{3}{c}{\textbf{K = 5 }} & \multirow{2}{*}{\textbf{mAP }} & \multicolumn{3}{c}{\textbf{K = 10 }} & \multicolumn{3}{c}{\textbf{K = 20 }} & \multirow{2}{*}{\textbf{mAP }} & \multirow{2}{*}{\textbf{WmAP }}\\
 &  &  & \textbf{P } & \textbf{R } & \textbf{F1 } & \textbf{P } & \textbf{R } & \textbf{F1 } &  & \textbf{P } & \textbf{R } & \textbf{F1 } & \textbf{P } & \textbf{R } & \textbf{F1 } &  &\\ 
\toprule[0.15em]

\multirow{2}{*}{LESA (M=10)} 
 &   & ZSL   & 25.7  & 41.1  & 31.6  & 19.7  & 52.5  & 28.7  & 19.4  & 0.7   & 25.6  & 1.4   & 0.5   & 37.4  & 1.0   & 41.7  & -\\
 &  & GZSL  & 23.6  & 10.4  & 14.4  & 19.8  & 14.6  & 16.8  & 5.6   & 16.2  & 18.9  & 17.4  & 10.2  & 23.9  & 14.3  & 45.4  & -\\ 
\cmidrule(lr){3-18}

\multirow{2}{*}{ZS-SDL} 
 & \multirow{2}{*}{ZS}  & ZSL   & 24.2  & 41.3  & 30.5  & 18.8  & 53.4  & 27.8  & 25.9  & 6.1   & 47.0  & 10.7  & 4.4   & 68.1  & 8.3   & 62.9  & -\\
 &  & GZSL  & 27.7  & 13.9  & 18.5  & 23.0  & 19.3  & 21.0  & 12.1  & 35.3  & 40.8  & 37.8  & 23.6  & 54.5  & 32.9  & 75.3  & -\\
\cmidrule(lr){3-18}

\multirow{2}{*}{BiAM$^*$} 
 &  & ZSL   & 26.6  & 42.5  & 32.7  & 20.5  & 54.6  & 29.8  & 25.9  & 3.9   & 30.7  & 7.0   & 2.7   & 41.9  & 5.5   & 65.6  & 72.9\\
 &  & GZSL  & 25.2  & 11.1  & 15.4  & 21.6  & 15.9  & 18.2  & 9.4   & 13.8  & 15.9  & 14.8  & 9.7   & 22.3  & 14.8  & \textbf{81.7}  & 85.0\\ 
\bottomrule[0.1em]

\multirow{2}{*}{CLIP-FT}

 & \multirow{5}{*}{OV} 
 & ZSL  & 19.1  & 30.5  & 23.5  & 14.9  & 39.7  & 21.7  & 30.5  & 10.8   & 84.0  & 19.1  & 5.9   & 92.1  & 11.1   & 66.2  & 88.2\\
 &  & GZSL  & 33.2  & 14.6  & 20.3  & 27.4  & 20.2  & 23.2  & 16.8  & 37.5  & 43.3  & 40.2  & 25.4  & \textbf{58.7}  & 35.4  & 77.5  & 85.9\\
\cmidrule(lr){3-18}

\multirow{2}{*}{MKT} 
 &  & ZSL   & \textbf{27.7}  & \textbf{44.3}  & \textbf{34.1}  & \textbf{21.4}  & \textbf{57.0}  & \textbf{31.1}  & \textbf{37.6}  
        & \textbf{11.1}  & \textbf{86.8}  & \textbf{19.7}  & \textbf{6.1}   & \textbf{94.7}  & \textbf{11.4}  & \textbf{68.1}   & \textbf{89.2} \\
 &  & GZSL  & \textbf{35.9}  & \textbf{15.8}  & \textbf{22.0}  & \textbf{29.9}  & \textbf{22.0}  & \textbf{25.4}  & \textbf{18.3}   
        & \textbf{37.8}  & \textbf{43.6}  & \textbf{40.5}  & \textbf{25.4}  & 58.5  & \textbf{35.4}  & 81.4   & \textbf{89.8} \\ 
\cmidrule(lr){3-18}
\bottomrule[0.1em]
\end{tabular}
}
\caption{
State-of-the-art comparison for ZSL and GZSL tasks on the NUS-WIDE and Open Images datasets. 
The results are reported in terms of mAP, as well as precision (P), recall (R), and F1 score at $K{\in}\{3,5\}$ for NUS-WIDE and $K{\in}\{10,20\}$ for Open Images.
`*' means that the results are reproduced based on official pre-trained models.
Bold indicates the best score.}
\label{tab:sota}
\end{table*}

\subsection{Prompt Tuning for Label Embedding}
Following~\cite{clip}, we first design a manual prompt template as ``There is a \{\textit{label}\} in the scene''. 
We fill up the blank in this template with label name and treat the whole sentence as the input of CLIP text encoder.
The output of CLIP text encoder is utilized as the label embedding.
Due to the different training objectives, we argue that the label embeddings generated by pre-trained CLIP text encoder are not optimal for multi-label classification.
Thus, we propose to further fine-tune the label embedding.
However, it is very hard to fine-tune the entire text encoder due to the mode collapse problem caused by insufficient training samples.
Motivated by CoOp~\cite{coop}, we introduce prompt tuning for the adaptation of label embedding.
During the tuning process, all parameters except for the context embedding of the prompt template, which illustrated as the dotted box in Figure~\ref{fig:overall_arch}, are fixed.
We show that compared with the hand-crafted prompt, continuous search in embedding space based on CLIP text encoder facilitates the learning of optimal context embedding for our task.

\subsection{Loss Functions}
We divide the training process of our method into two stages. 
In the first stage, label embedding is generated by the pre-trained CLIP text encoder, and the vision encoder is trained with the objectives of ranking loss and distillation loss,
\begin{equation}
\mathcal{L}_{\text {stage1}} = \mathcal{L}_{\text {rank }} + \lambda \mathcal{L}_{dist},
\end{equation}
where $\lambda$ is the weight factor of knowledge distillation. \par
In the second stage, we only finetune the context embedding with the objective of ranking loss,
\begin{equation}
\mathcal{L}_{\text {stage2}} = \mathcal{L}_{\text {rank }}.
\end{equation}

\section{Experiments}

\subsection{Experiments Setup}
\noindent\textbf{Datasets:} 
In the \textbf{NUS-WIDE} dataset, there are 81 human verified labels, in addition to 925 labels based on Flickr user tags. 
Similar to LESA~\cite{lesa}, we treat 925 labels as seen labels and the other 81 labels as unseen labels.
Following official train/test split, we utilize 161,789 images for training and 107,859 images for testing.
The \textbf{Open Images} (v4) dataset is more challenging because it consists of 9M training images and 125,456 testing images. 
Similar to LESA, we treat 7,186 labels with more than 100 images in training set as seen and the most frequent 400 test labels that are not present in training data as unseen. \\ 
\noindent\textbf{Metrics:} 
Following LESA, we use mean Average Precision (mAP) and F1 score at \textit{top}-$K$ predictions to evaluate our method.
The mAP reflects the ranking accuracy of each label across all images and the F1 score reflects the label ranking accuracy of each image. \\ 
\noindent\textbf{Implementation Details:} 
We use the ImageNet-1K pre-trained ViT-B/16 as our vision backbone.
As for the two-stream module, the local head consists of two linear layers, and the global head is a linear projection layer.
To generate label embedding and conduct knowledge distillation on vision encoder, we select the pre-trained CLIP with ViT-B/16 image encoder as our VLP model.
Patch projection of ViT-B/16 yields $14 \times 14=196$ patches for an image with a resolution of $224 \times 224$.
The $k$ for \textit{top}-$k$ pooling is set to 18, and the weight of knowledge distillation $\lambda$ is set to 1.
In the first stage, we use AdamW optimizer with base learning rate of $0.001$ and weight decay of $0.005$.
We adjust base learning rate of the AdamW optimizer to $0.00003$ during the second stage for fine-tuning the context embedding.
On NUS-WIDE, we train the model for 20 epochs with the mini-batch of 128 and 10 epochs with the mini-batch of 16 in the first and second stage, respectively.
Considering the large scale of Open Images, the model is trained for 4 epochs and 2 epochs in each stage with the same batch size as above.

\subsection{State-of-the-art Comparison}

In this experiment, we compare our model with traditional ML-ZSL methods. 
Also, we fine-tune the pre-trained CLIP on base categories with ranking loss and denote it as CLIP-FT. 
As a new OV-ML baseline, CLIP-FT surpasses most existing ML-ZSL methods on mAP.
The experimental results on zero-shot learning(ZSL) and generalized zero-shot learning(GZSL) tasks are shown in Table~\ref{tab:sota}. 
The mAP and F1 scores at \textit{top}-$K$ ($K\in\{3,5\}$ for NUS-WIDE and
$K\in\{10,20\}$ for Open Images) are reported. 

On NUS-WIDE, the recently proposed BiAM~\cite{biam}, which utilizes a bi-level attention module to enrich the features, acquires the best results in ZSL task with mAP score of 25.9\%. 
MKT surpasses BiAM with an absolute gain of 11.7\% mAP and improves the F1 score by absolute gains of 1.4\% and 1.3\% at $K {=} 3$ and $K {=} 5$, respectively. 
In GZSL task, the approach of ZS-SDL~\cite{zssdl} achieves the best scores with 12.1\% mAP. 
MKT improves the mAP by an absolute gain of 6.5\% and reaches state of the art in terms of F1 score with 22.0\% at $K {=} 3$ and 25.4\% at $K {=} 5$.
Compared with CLIP-FT, MKT shows significant improvement on both ZSL and GZSL task.

On Open Images, following BiAM, we also calculate mAP weighted on different sample numbers(denoted as WmAP).
ZS-SDL reaches the state of the art before in terms of F1 score in both ZSL and GZSL tasks. 
MKT achieves consistent improvement over it with absolute gains of 9.0\%/2.7\% and 3.1\%/2.5\% at $K {=} 10$ and $K {=} 20$ on ZSL/GZSL task. 
In comparison with previous best results on mAP/WmAP metric, MKT outperforms BiAM by 2.5\%/16.3\% on ZSL and have a comparable performance on GZSL task. 
MKT also surpasses CLIP-FT on both tasks.

\begin{table}[t]
\centering
\setlength{\tabcolsep}{6pt}

\adjustbox{width=1\linewidth}{
\begin{tabular}{cc|cccc} 
\toprule[0.15em]
\multirow{1}{*}{\textbf{Distill}} & \multirow{1}{*}{\textbf{Prompt}} & \multirow{1}{*}{\textbf{Task }} & \multirow{1}{*}{\textbf{mAP }} & \multicolumn{1}{c} {\textbf{F1 (K = 3) }} & \multicolumn{1}{c}{\textbf{F1 (K = 5) }}   \\
\toprule[0.15em]
\multirow{2}{*}{\XSolidBrush} &
\multirow{2}{*}{\XSolidBrush}  
& ZSL   & 32.4  & 29.4  & 26.5  \\
&& GZSL  & 16.8  & 21.0  & 24.0  \\ 
\cmidrule(lr){3-6}

\multirow{2}{*}{\Checkmark} &
\multirow{2}{*}{\XSolidBrush}
& ZSL   & 37.3  & 32.5  & 29.5  \\
&& GZSL  & 18.2  & 21.7  & 24.9  \\ 
\cmidrule(lr){3-6}

\multirow{2}{*}{\XSolidBrush} &
\multirow{2}{*}{\Checkmark} 
& ZSL   & 32.5  & 29.5  & 26.4 \\
&& GZSL  & 16.8  & 21.1  & 24.1  \\ 
\cmidrule(lr){3-6}

\multirow{2}{*}{\Checkmark} &
\multirow{2}{*}{\Checkmark} 
& ZSL   & \textbf{37.6}  & \textbf{34.1}  & \textbf{31.1} \\
&& GZSL  & \textbf{18.3}  & \textbf{22.0}  & \textbf{25.4}  \\ 
\cmidrule(lr){3-6}

\bottomrule[0.1em]
\end{tabular}
}
\caption{Impact of knowledge distillation and prompt tuning.}
\label{tab:ablation}
\end{table}
\begin{table}[ht]
\centering
\setlength{\tabcolsep}{6pt}
\adjustbox{width=1\linewidth}{
\begin{tabular}{ccccc} 
\toprule[0.15em]
\multirow{1}{*}{\textbf{Embedding}} & \multirow{1}{*}{\textbf{Task }} & \multirow{1}{*}{\textbf{mAP }} & \multicolumn{1}{c} {\textbf{F1 (K = 3) }} & \multicolumn{1}{c}{\textbf{F1 (K = 5) }}   \\
\toprule[0.15em]
\multirow{2}{*}{GloVe} 
& ZSL   & 27.1  & 22.8  & 21.4  \\
& GZSL  & 16.1  & 20.6  & 23.4  \\ 
\cmidrule(lr){2-5}
\multirow{2}{*}{CLIP}           
& ZSL   & \textbf{32.4}  & \textbf{29.4}  & \textbf{26.5}  \\
& GZSL  & \textbf{16.8}  & \textbf{21.0}  & \textbf{24.0}  \\ 
\bottomrule[0.1em]
\end{tabular}
}
\caption{Impact of label embedding. 
For a fair comparison, we only change label embedding and train both models without knowledge distillation or prompt tuning.}
\label{tab:glove}
\end{table} 
\begin{figure}[ht]
  \centering
  \subcaptionbox{Top-3 Accuracy\label{fig:top3acc}}
    {\includegraphics[width=0.32\linewidth, page=3]{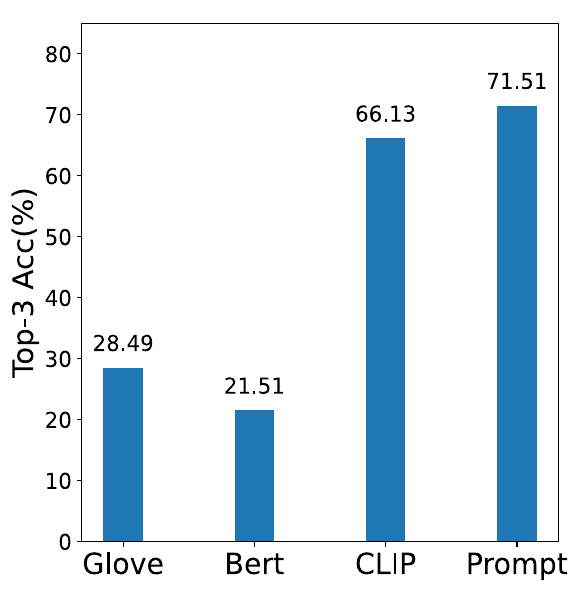}}
  \subcaptionbox{Top-3 Retrieved Results\label{fig:label_ret}}
    {\includegraphics[width=0.6\linewidth, page=3]{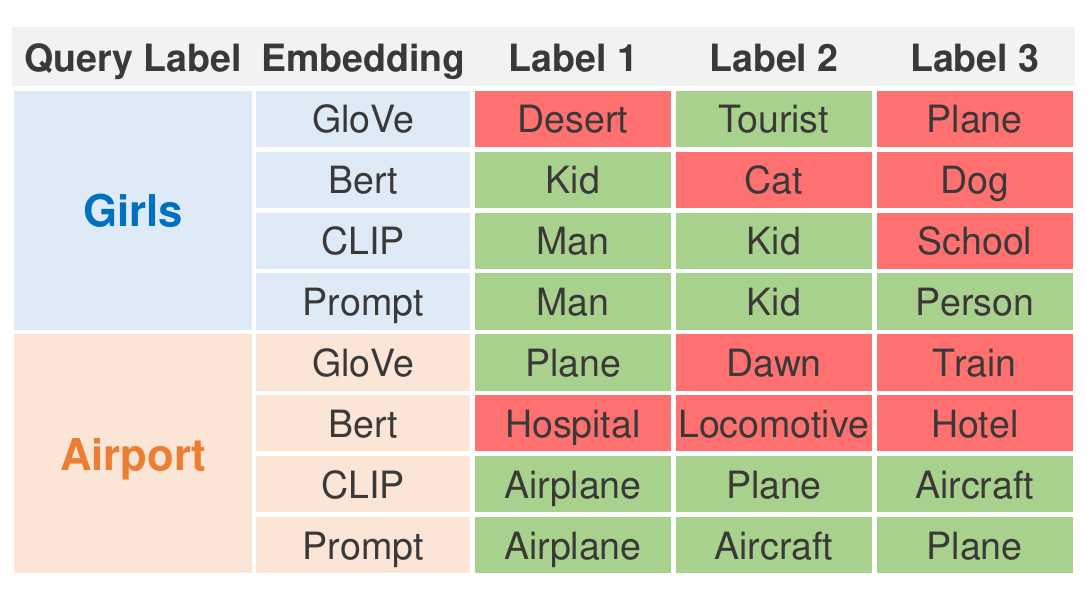}}
  \caption{Results of label retrieval. 
 Overall Top-3 accuracy and examples of retrieved labels are reported. Retrieved labels belonging to the same major category with the query label are considered to be correct (in green).
 }
  \label{fig:label_result}
\end{figure}

\subsection{Ablation Studies}
\noindent\textbf{Effects of knowledge distillation and prompt tuning:}
To study the impacts of knowledge distillation and prompt tuning, 
we conduct experiments with different training schemes and illustrate the results in Table~\ref{tab:ablation}.
We take the first row as the baseline for the following comparisons, which is trained without knowledge distillation and prompt tuning.
It shows that the introduction of knowledge distillation improves the performance on both ZSL and GZSL tasks.
We conjecture that knowledge distillation not only facilitates the image embedding to align better with VLP model based label embedding but also suppresses the overfitting of the model to seen labels.
Moreover, we observe that prompt tuning can further improve performance.
It can be attributed to the reason that the prompt-tuned context embedding tends to pay more attention to the visual information that benefits image classification.
Compared with the baseline in the first row, MKT shows significant improvement with the combination of knowledge distillation and prompt tuning.

\noindent\textbf{Comparison of label embedding:}
Because prediction results are based on the similarity between image and label embeddings, label embedding has a significant impact on model performance.
Table~\ref{tab:glove} shows the results of baseline model with VLP model based and GloVe based label embeddings.
Compared with the model based on GloVe embedding, the VLP embedding based model achieves superior performance on both ZSL and GZSL task.
We speculate that language models like GloVe or Bert cannot capture visual consistency among similar labels because of the lack of visual information during the training process, thus limiting the generalization ability to unseen labels.
To validate our assumption, we conduct a label retrieval experiment.
We select 62 common labels in NUS-WIDE and divide them into 14 major categories based on their visual and semantic similarity. 
Both language models (\ie, GloVe and Bert) and VLP models (\ie, CLIP and its prompt-tuned version) are utilized to generate label embeddings.
All embeddings are normalized, and cosine similarity is used to retrieve the most similar embeddings.
Figure~\ref{fig:label_result} illustrates the retrieval results with the overall Top-3 accuracy and examples of retrieved labels.
Notice that compared with language model, VLP model can capture both semantic and visual consistency between labels.
For instance, ``\textit{girls}'' contains similar visual information with its retrieved labels ``\textit{man}'', ``\textit{kid}'' and ``\textit{person}''. 
We argue that label embedding with both visual and semantic consistency facilitates the generalization to unseen labels.

\begin{figure}[t]
  \centering
  \subcaptionbox{Global Head Prediction\label{fig:global-head-score}}
    {\includegraphics[width=0.49\linewidth]{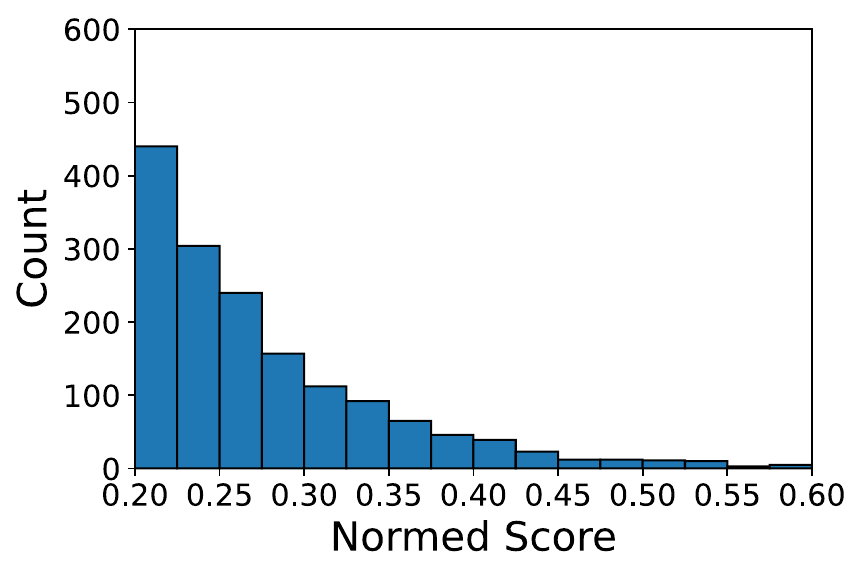}}
  \subcaptionbox{Local Head Prediction\label{fig:local-head-score}}
    {\includegraphics[width=0.49\linewidth]{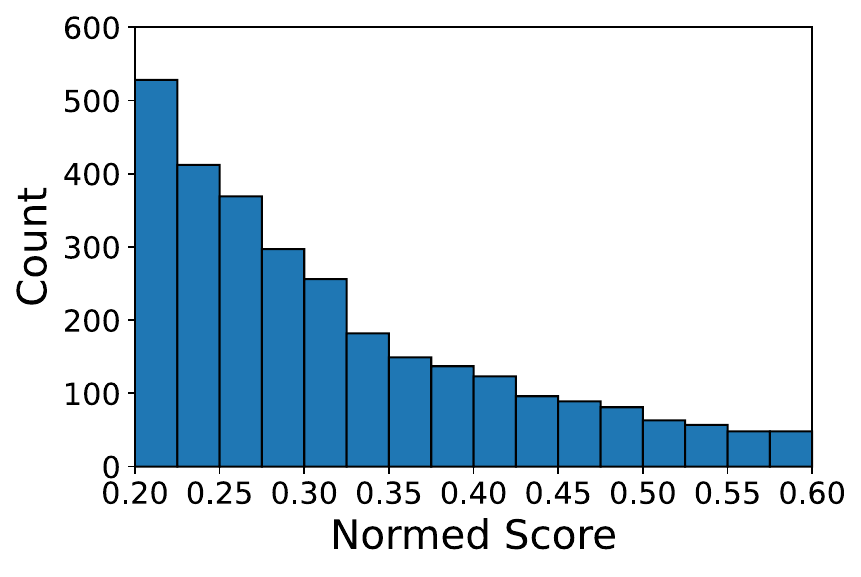}}
  \caption{Distribution of global and local predictions.}
  \label{fig:global_local}
\end{figure}

\begin{table}[t]
\centering

\setlength{\tabcolsep}{6pt}
\adjustbox{width=1\linewidth}{
\begin{tabular}{cc|cccc} 
\toprule[0.15em]
\multirow{1}{*}{\textbf{Local}} & \multirow{1}{*}{\textbf{Global}} & \multirow{1}{*}{\textbf{Task}} & \multirow{1}{*}{\textbf{mAP}} & \multicolumn{1}{c} {\textbf{F1 (K = 3) }} & \multicolumn{1}{c}{\textbf{F1 (K = 5) }}   \\
\toprule[0.15em]
\multirow{2}{*}{\Checkmark} &
\multirow{2}{*}{\XSolidBrush}
& ZSL   & 29.1  & \textbf{29.9}  & \textbf{27.2}  \\
& & GZSL  & \underline{15.7}  & \underline{20.8}  & \underline{23.8}  \\ 
\cmidrule(lr){3-6}
\multirow{2}{*}{\XSolidBrush} &
\multirow{2}{*}{\Checkmark}  
& ZSL   & \underline{30.3}  & 23.3  & 21.4  \\
& & GZSL  & 15.5  & 19.4  & 22.1  \\ 
\cmidrule(lr){3-6}
\multirow{2}{*}{\Checkmark} &
\multirow{2}{*}{\Checkmark}           
& ZSL   & \textbf{32.4}  & \underline{29.4}  & \underline{26.5}  \\
& & GZSL  & \textbf{16.8}  & \textbf{21.0}  & \textbf{24.0}  \\ 
\cmidrule(lr){3-6}
\bottomrule[0.1em]
\end{tabular}
}
\caption{Effectiveness of the two-steam module. 
Bold indicates the best, and underline indicates the second best.
}
\label{tab:head}
\end{table}

\noindent\textbf{Effect of the two-stream module:}
To demonstrate the effectiveness of our proposed two-stream module, we conduct ablation studies of both local and global heads.
Table \ref{tab:head} shows the results in terms of mAP and F1 score on NUS-WIDE.
Notice that the global head only model performs well on mAP while the local head only model achieves better F1 score in ZSL task.
We speculate that this is due to the fact that the global representation is more general while the local representation is more discriminative.
As illustrated in Figure \ref{fig:global_local}, the local head tends to predict higher scores than the global head.
While the more discriminative feature allows relevant labels to stand out, it also makes the model more sensitive to noise, leading to wrong predictions.
On the other hand, compared to F1 score, mAP is more susceptible to the wrong predictions with high scores. 
Therefore, the local head only model acquires better F1 score and inferior mAP.
With the combination of local and global heads, the two-stream module can acquire more discriminative predictions with resistance to noise, leading to higher performance.
\begin{figure}[t]
\centering
\subcaptionbox{Variation of $\lambda$\label{fig:distw}}
    {\includegraphics[width=0.49\linewidth, page=1]{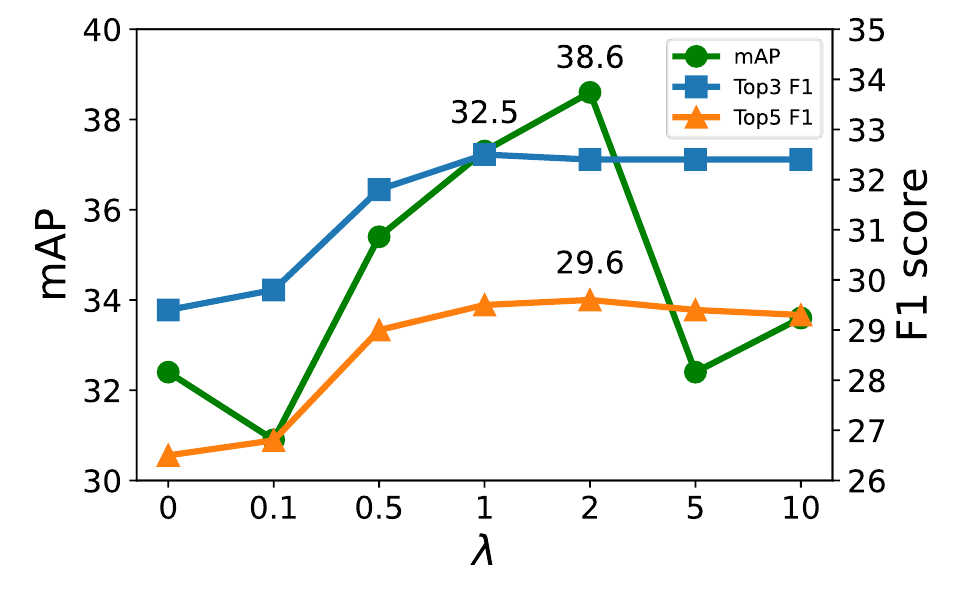}}
  \subcaptionbox{Variation of $k$\label{fig:head_num}}
    {\includegraphics[width=0.49\linewidth, page=1]{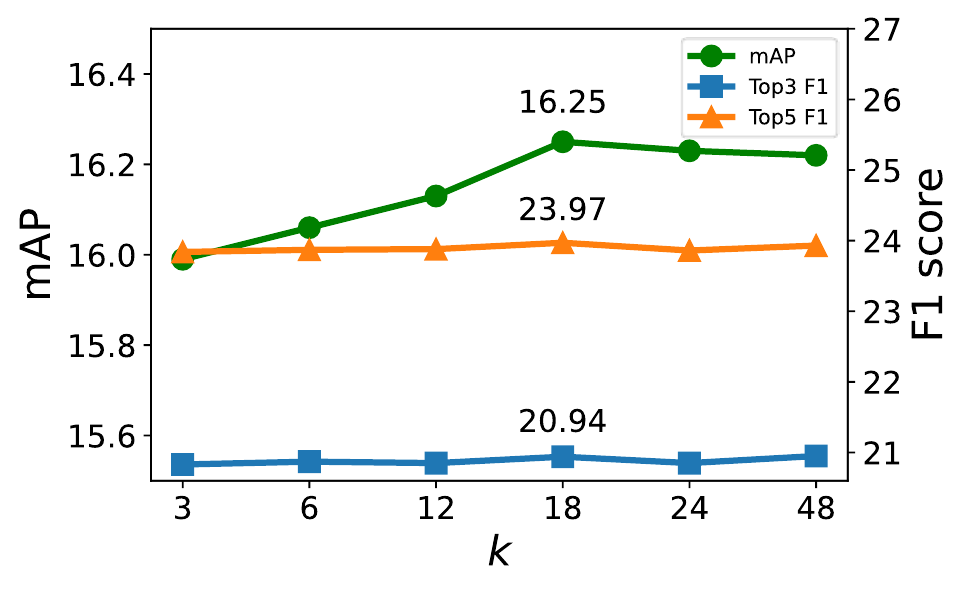}}
\caption{Impact of hyper-parameters. 
The results of ZSL task with respect to distillation weight $\lambda$ and GZSL task with respect to $k$ for \textit{top}-$k$ in local head are presented.
}
\label{fig:hyper}
\end{figure}
\begin{figure}[ht]
\centering
\includegraphics[width=1\columnwidth, page=5]{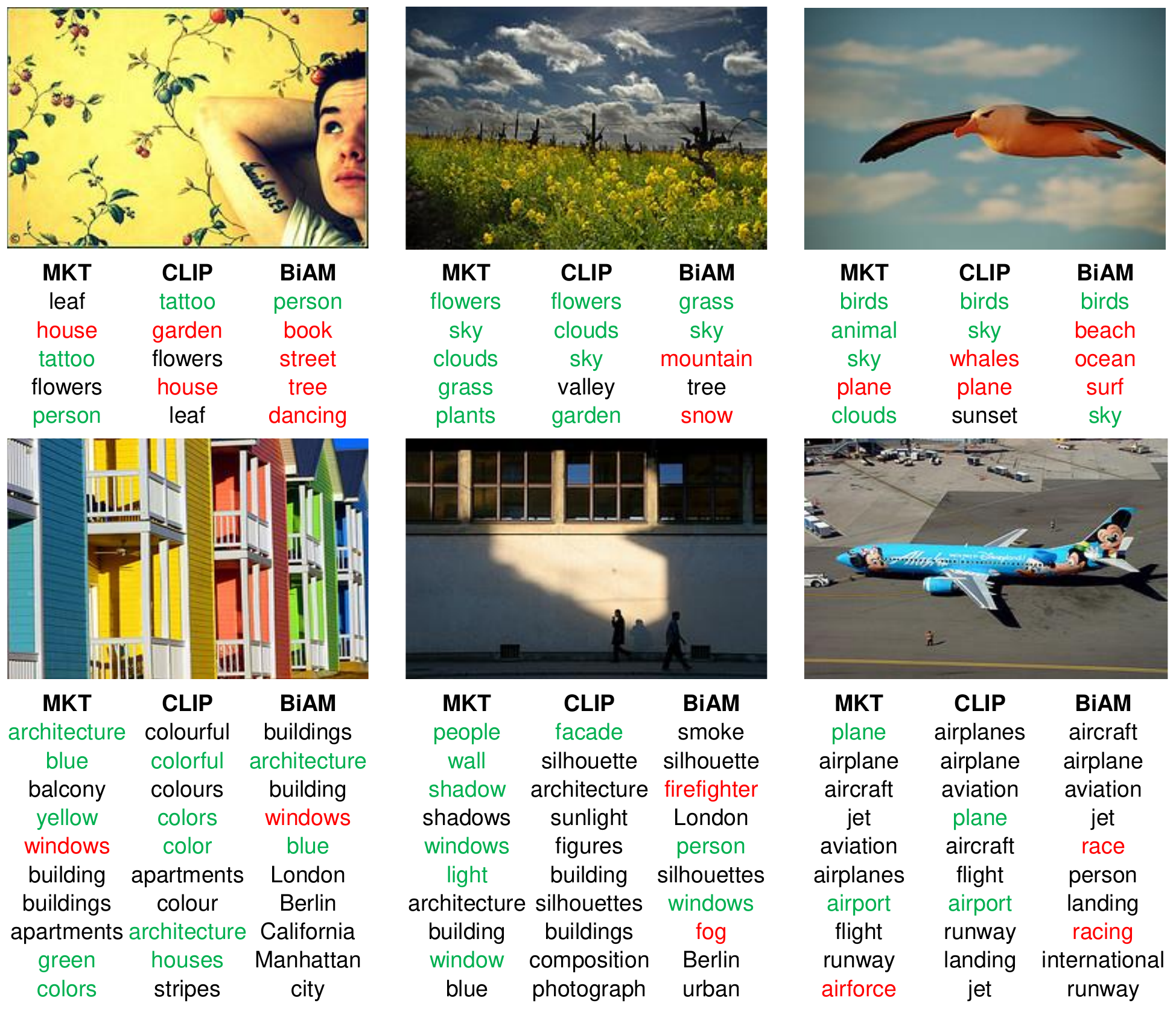}
\caption{
Comparison of predictions on test samples from NUS-WIDE. 
The top row shows the Top-5 prediction in ZSL task, and the bottom is the Top-10 prediction in GZSL task. 
True positive predictions are shown in green and the red font denotes apparently incorrect predictions. }
\label{fig:prediction}
\end{figure}
\noindent\textbf{Varying the hyper-parameters:}
Here, we explore the effect of knowledge distillation and variation of $k$ value in the local head.
Because knowledge distillation aims to transfer zero-shot classification ability, we are more concerned about its performance on unseen labels.
Figure~\ref{fig:distw} illustrates the results of ZSL task with respect to distillation weight $\lambda$. 
Notice that when $\lambda$ is smaller than 1, the performance of our approach improves because knowledge distillation facilitates the alignment of image and label embeddings. 
However, there is a drop in performance when $\lambda$ is larger than 2. 
We argue that too large $\lambda$ may impair the learning of classification objective $\mathcal{L}_{\text {rank }}$.
The two-stream module is designed to improve the recognition of multiple labels, so we focus more on its ability to classify all labels.
Figure~\ref{fig:head_num} illustrates the results of GZSL when altering $k$ value in the local head.
As $k$ increases, F1 score reaches the highest when $k {=} 18$.
We argue that when $k$ is too small, the local head output is sensitive to noise. 
On the other hand, if $k$ is too large, the output will be less discriminative. 
For example, if $k$ is set as the total patch number, \textit{top}-$k$ pooling will be equal to global average pooling.
In contrast to F1 score, mAP tends to increase while $k$ value increases.
When $k$ is small, the local head output tends to be discriminative but sensitive to noise, resulting in a lower mAP value.
As $k$ increases, the output becomes moderate and more resistant to noise, leading to a higher mAP value.
\begin{figure}[t]
\centering
\includegraphics[width=0.9\columnwidth, page=2]{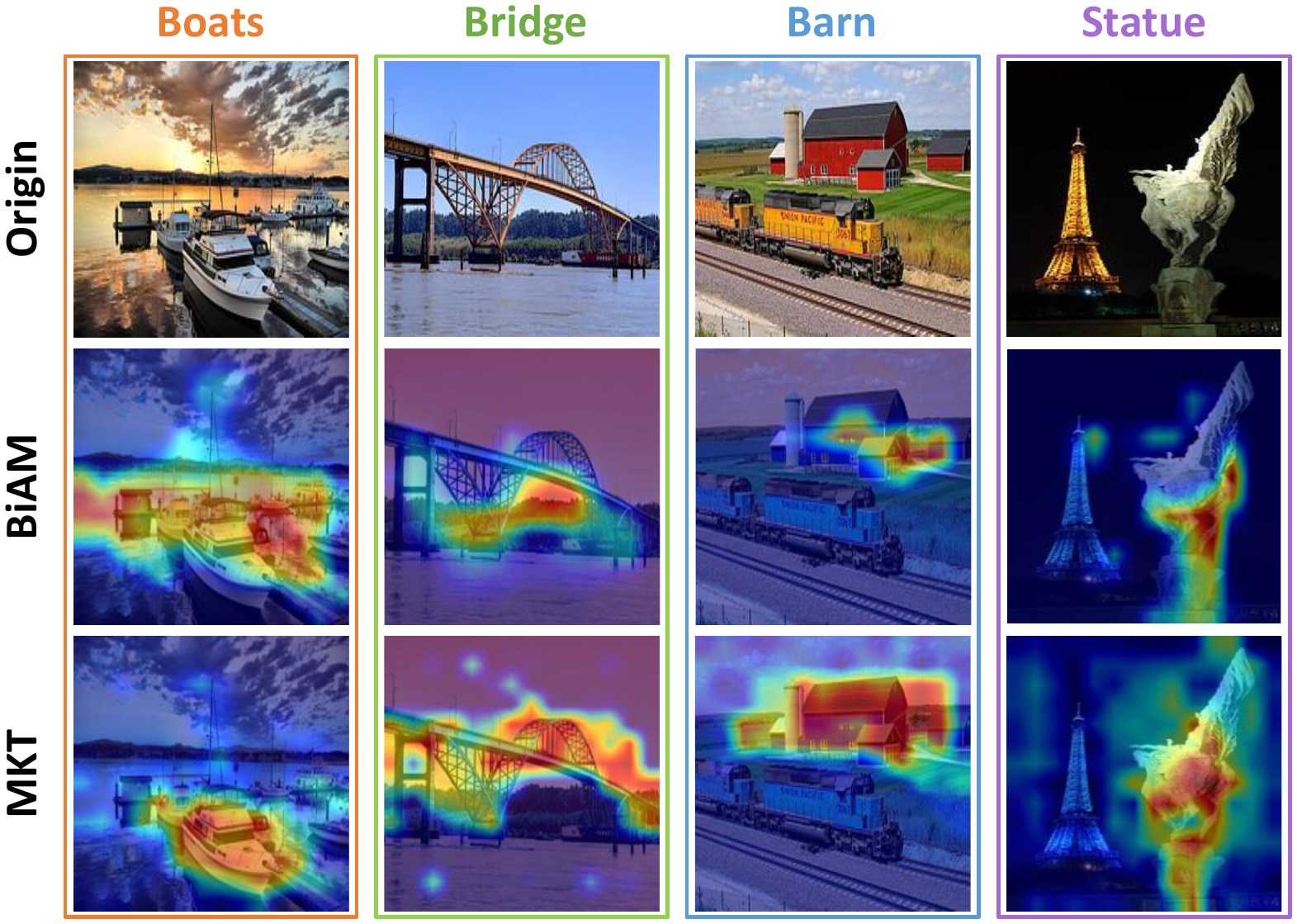}
\caption{
Comparison of Grad-CAM visualization.
}
\label{fig:heatmap}
\end{figure}
\subsection{Qualitative Assessment}
In this section, we visualize both predictions and attention maps on several samples.
Figure~\ref{fig:prediction} presents predictions of CLIP, BiAM and our approach on ZSL and GZSL tasks respectively. 
Compared with CLIP, our approach produces more diverse predictions because the two-stream module captures discriminative features. 
Compared to BiAM, our model with VLP based label embedding can identify semantic and visual similarity among labels.  
For example, in the last sample of Figure~\ref{fig:prediction},  label ``\textit{plane}'', ``\textit{airplane}'' and ``\textit{aircraft}'' are synonymous and should have similar scores.
Figure~\ref{fig:heatmap} illustrates the comparison of attention maps between BiAM and ours. 
The results show that our method can capture relevant regions more precisely. 
For instance, in the first column, BiAM pays attention to large irrelevant areas while our method exactly focuses on the boat region.

\section{Conclusion}
In this work, we propose an open-vocabulary based multi-modal knowledge transfer (MKT) framework for multi-label classification, which jointly exploits semantic multi-modal information in image-text pairs based VLP models.
To facilitate transferring the image-text matching ability of VLP model to classification, knowledge distillation and prompt tuning are introduced.
Additionally a two-stream module is proposed to capture both local and global features, leading to significant performance gains in multi-label tasks.
Extensive results demonstrate that our model surpasses previous ML-ZSL methods and establishes a new state of the art for open-vocabulary multi-label classification on NUS-WIDE and Open Images datasets.
This is the first work in open-vocabulary multi-label classification and it is expected to encourage future works to explore multi-modal knowledge applications in classification.
\section{Acknowledgements}
This work is supported in part by the National Natural Science Foundation of China under Grant 62171248, the Natural Science Foundation of Guangdong Province 2021A1515011807, Shenzhen Science and Technology Program (Grant No.RCYX20200714114523079, JCYJ20220818101012025) and the PCNL KEY project (PCL2021A07), and Shenzhen Science and Technology Innovation Commission (Research Center for Computer Network (Shenzhen) Ministry of Education).
\bibliography{aaai23}

\clearpage
\appendix
\begin{center} \begin{Large}{\textbf{Supplementary Material}} \end{Large} 
\vspace{10pt}
\end{center}
In this Supplementary Material, we present additional details and analyses to further understand our open-vocabulary multi-label classification method MKT.
This material is organized as follows:
\begin{itemize}
    \item Comparison with open vocabulary methods.
    \item Experiment on standard multi-label classification.
    \item Robustness to backbone variants.
    \item Text-Label classification experiment.
    \item Details of label selection and t-SNE visualization of different label embeddings.
    \item Definition and discussion of evaluation metrics. 
    \item Comparison of complexity. 
    \item Additional analysis about knowledge distillation. 
    \item Additional analysis about prompt tuning.
    \item Additional qualitative prediction results. 
\end{itemize}

\section{Comparison with Open Vocabulary Methods}
In this section, we compare our model with recently proposed open vocabulary multi-label classification methods (\eg, DualCoOp~\cite{sun2022dualcoop} and ADDS~\cite{xu2022dual}). 
Results on NUS-WIDE dataset are shown in Table~\ref{tab:ov}.
For a fair comparison, we also initialize our vision backbone with the CLIP pretrained model and denote it as MKT (CLIP).
As shown in Table~\ref{tab:ov}, though DualCoOp achieves higher performance on ZSL task, MKT shows significant improvements on GZSL task. 
Compared with ZSL task, GZSL task is more challenging because the model has to recognize both seen and unseen labels, and more practical because the seen labels are usually more common in practice.
\begin{table}[h]
\centering
\caption{Comparison with other open vocabulary methods on NUS-WIDE. Results are reported in terms of mAP and F1 score. 
}
\footnotesize
\begin{tabular}{ccccc} 
\toprule[0.15em]
\multirow{1}{*}{\textbf{Method}} & \multirow{1}{*}{\textbf{Task }} & \multirow{1}{*}{\textbf{mAP }} & \multicolumn{1}{c} {\textbf{F1 (K = 3) }} & \multicolumn{1}{c}{\textbf{F1 (K = 5) }}   \\
\toprule[0.15em]
\cmidrule(lr){2-5}
\multirow{2}{*}{DualCoOp}           
& ZSL   & \textbf{43.6}  & \textbf{41.3}  & \textbf{38.7} \\
& GZSL  & 12.0  & 19.4  & 22.1  \\ 
\cmidrule(lr){2-5}
\multirow{2}{*}{ADDS (224)}           
& ZSL   & 36.6  & 34.2  & 36.6 \\
& GZSL  & -  & -  & -  \\ 
\cmidrule(lr){2-5}
\multirow{2}{*}{ADDS (336)}           
& ZSL   & 39.0  & \underline{37.0}  & \underline{39.3} \\
& GZSL  & -  & -  & -  \\ 
\cmidrule(lr){2-5}
\multirow{2}{*}{MKT (IM-1K)}           
& ZSL   & 37.6  & 34.1  & 31.1 \\
& GZSL  & \underline{18.3}  & \underline{22.0}  & \underline{25.4}  \\ 
\cmidrule(lr){2-5}
\multirow{2}{*}{MKT (CLIP)}           
& ZSL   & \underline{42.5}  & 34.1  & 31.5  \\
& GZSL  & \textbf{21.5}  & \textbf{24.1}  & \textbf{27.9}  \\ 
\bottomrule[0.1em]
\end{tabular}
\label{tab:ov}
\end{table}

\section{Standard Multi-label Classification}
\label{standard}
In addition to traditional multi-label zero-shot learning setting, we also conduct a standard multi-label classification experiment to validate the multi-label recognition ability of MKT.
In this setting, all labels are available during the training stage.
Concretely, following previous work~\cite{biam}, we select 81 human-annotated labels from NUS-WIDE dataset for both training and evaluating.
As shown in Table~\ref{tab:standard}, our method MKT outperforms existing ML-ZSL methods by a large margin. 
It can be attributed to both the efficient design of MKT and multi-modal label embedding which is more suitable for vision tasks.
\begin{table}[t]
\centering
\caption{Comparison for the standard multi-label classification on NUS-WIDE. Results are reported in terms of mAP, as well as precision (P), recall (R), and F1 score. 
}
\adjustbox{width=1\linewidth}{
\begin{tabular}{cccccccc} 
\toprule[0.15em]
\multirow{2}{*}{\textbf{ Method}} 
& \multicolumn{3}{c}{\textbf{K = 3 }} & \multicolumn{3}{c}{\textbf{K = 5 }} & \multirow{2}{*}{\textbf{mAP }} \\
& \textbf{P } & \textbf{R } & \textbf{F1 } & \textbf{P } & \textbf{R } & \textbf{F1 } &  \\ 
\toprule[0.15em]
Logistic
& 46.1  & 57.3  & 51.1  & 34.2  & 70.8  & 46.1  & 21.6  \\
\cmidrule(lr){2-8}
WARP
& 49.1  & 61.0  & 54.4  & 36.6  & 75.9  & 49.4  & 3.1  \\
\cmidrule(lr){2-8}
WSABIE
& 48.5  & 60.4  & 53.8  & 36.5  & 75.6  & 49.2  & 3.1  \\
\cmidrule(lr){2-8}
Fast0Tag
& 48.6  & 60.4  & 53.8  & 36.0  & 74.6  & 48.6  & 22.4  \\
\cmidrule(lr){2-8}
CNN-RNN
& 49.9  & 61.7  & 55.2  & 37.7  & 78.1  & 50.8  & 28.3  \\
\cmidrule(lr){2-8}
LESA
& 52.3  & 65.1  & 58.0  & 38.6  & 80.0  & 52.0  & 31.5  \\
\cmidrule(lr){2-8}
GAN-MLZSL
& 53.5  & 66.5  & 59.3  & 39.4  & 81.6  & 53.1  & 46.7  \\
\cmidrule(lr){2-8}
BiAM
& -     & -     & 59.6  & -     & -     & 53.4  & 47.8  \\
\cmidrule(lr){2-8}
MKT 
& \textbf{56.5}  & \textbf{70.2}  & \textbf{62.6}  
& \textbf{41.3}  & \textbf{85.6}  & \textbf{55.7}  & \textbf{60.3}  \\
\bottomrule[0.1em]
\end{tabular}
}
\label{tab:standard}
\end{table}
\begin{table}[t]
\centering
\footnotesize
\caption{Robustness to Backbone Variants. We modified MKT with VGG-19 and DINO ResNet-50 backbones. Results are reported in terms of mAP on both benchmarks. 
}
\begin{tabular}{cccc|cc} 
\toprule[0.15em]
\multirow{2}{*}{\textbf{Backbone}} & \multirow{2}{*}{\textbf{Method}} &
\multicolumn{2}{c}{\textbf{NUS-WIDE}} &  \multicolumn{2}{c}{\textbf{Open Images}} \\
& & ZSL & GZSL & ZSL & GZSL \\
\toprule[0.15em]
\multirow{3}{*}{VGG-19} 
& LESA & 19.4 & 5.6 & 41.7 & 45.4\\
& BiAM & 26.3 & 9.3 & 73.6 & 84.5\\
& MKT & \textbf{28.5} & \textbf{9.9}  
& \textbf{87.0} 
& \textbf{87.1}\\
\hline
\multirow{3}{*}{ResNet-50} 
& LESA & 20.5 & 6.4 & 41.9 & 45.5 \\
& BiAM & 27.4 & 10.2 & 74.0 & 84.8\\
& MKT & \textbf{31.5} & \textbf{14.9}
& \textbf{87.7}
& \textbf{85.7}\\
\bottomrule[0.1em]
\end{tabular}
\label{tab:switch_backbone}
\end{table}

\section{Robustness to Backbone Variants}
\label{robustness}
For a fair comparison with existing ML-ZSL works such as LESA~\cite{lesa} and BiAM~\cite{biam}, we explore MKT performance with the same backbone (\eg, VGG19~\cite{vgg} and DINO ResNet-50~\cite{dino}) to exclude impact of superior backbones. 
In these CNN backbones, we treat the feature of the last convolution layer before global pooling as the local feature and the feature of the last fully connected layer as the global feature.
A similar two-stream module is employed on these features.
Considering the CNN backbone, we select CLIP with ResNet-101 backbone as the VLP model.
Table~\ref{tab:switch_backbone} shows that our method MKT surpasses LESA~\cite{lesa} and BiAM~\cite{biam} with the same backbone on both NUS-WIDE~\cite{nuswide} and Open Images~\cite{openiamge}.
Meanwhile, it also demonstrates that the design of MKT can easily be transferred to various backbones and boost their performance consistently.

\begin{table}[ht]
\centering
\footnotesize
\caption{Comparison on Text label. We evaluate both BiAM and MKT on 3756 text labels from Open Images dataset. Results are reported in terms of mAP and F1 scores. 
}
\begin{tabular}{cc|ccc} 
\toprule[0.15em]
\textbf{Method} & \textbf{Task} & \textbf{mAP} & \textbf{F1 (K = 10) } & \textbf{F1 (K = 20) } \\
\toprule[0.15em]
\multirow{2}{*}{BiAM} 
& ZSL & 69.70 & 4.5 & 3.7 \\
& GZSL & \textbf{78.76} & \textbf{12.4} & \textbf{9.42} \\
\hline
\multirow{2}{*}{MKT}
& ZSL & 79.04 & 22.6 & 12.4 \\
& GZSL & \textbf{82.52} & \textbf{31.5} & \textbf{22.9} \\
\bottomrule[0.1em]
\end{tabular}
\label{tab:text}
\end{table}
\begin{table}[ht]
\caption{Label list. We select 62 labels from both seen and unseen label sets, and divide them into 14 major categories based on their semantic and visual similarity.}
\adjustbox{width=\linewidth}{
\begin{tabular}{cl}
\toprule[0.15em]
Major Category & \multicolumn{1}{c}{Labels} \\ 
\hline
Human & kid, girls, mother, man, tourist, person \\
Animal & horse, bear, elk, dog, cow, zebra, fox, tiger, horses, cat \\
Plant & plant, flowers, tree, grass, garden \\
Place & school, hotel, hospital, chapel, restaurant \\
Terrestrial & mountains, desert, glacier, clouds, mountain \\
Water & river, ocean, water, waterfall, lake \\
Time & sunrise, dawn, darkness, sunset, nighttime \\
Vehicle & truck, vehicle, cars \\
Building & buildings, cityscape, tower, temple \\
Rail & locomotive, railroad, train \\
Flight & aircraft, airplane, airport, plane \\
Ship & ship, boats \\
Material & wood, stone, metal \\
Furniture & carpet, chairs \\ 
\toprule[0.15em]
\label{tab:labellist}
\end{tabular}}
\end{table}
\begin{figure*}[ht]
    \centering
    \includegraphics[width=0.9\textwidth, page=3]{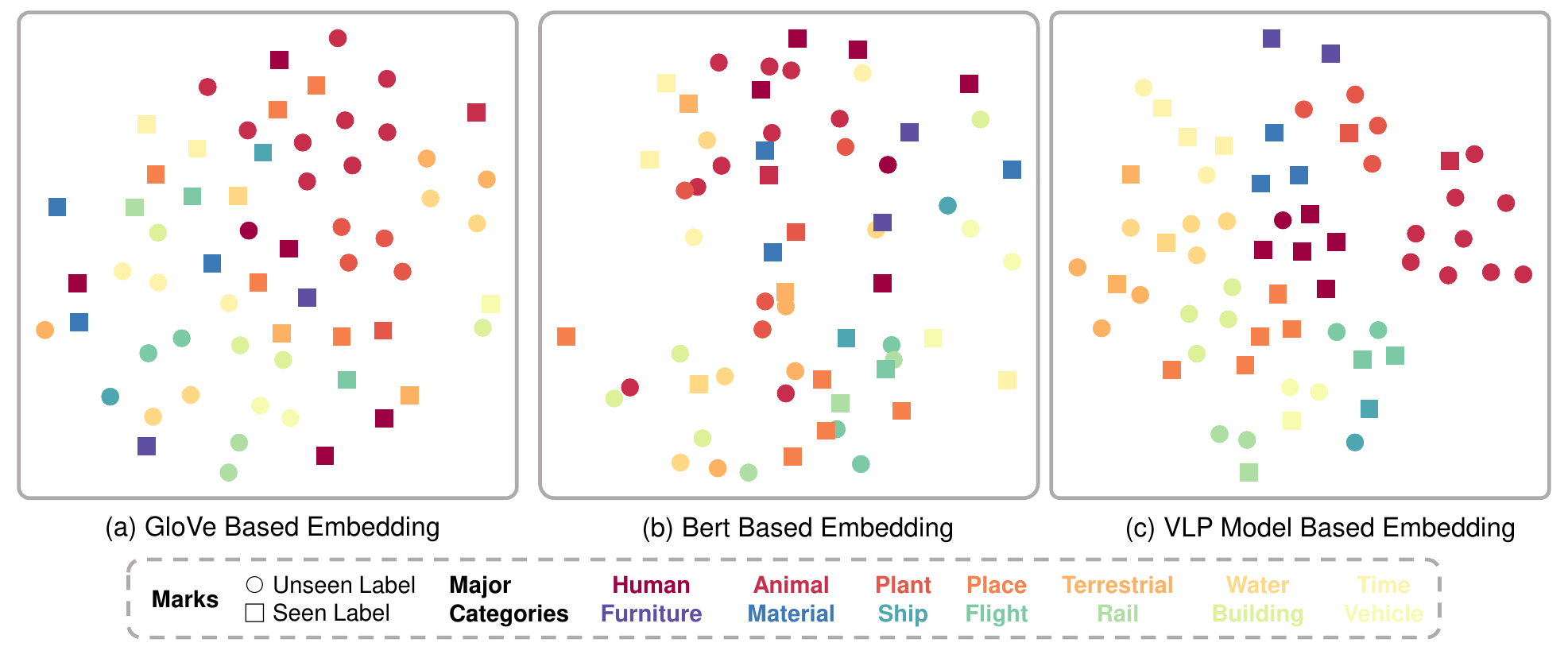}
\caption{
t-SNE visualization of label embeddings based on different models.
Same color denotes the labels (\eg, \textit{kid}, \textit{woman}) that belong to same major category like \textbf{Human} (in dark red).  
For Bert and VLP model, label embeddings are generated with the same manual template ``There is a \{label\} in the scene''. Compared with GloVe and Bert, VLP model can capture semantic and visual consistency among similar labels. (Best viewed in color.) 
    }
    \label{fig:labelemb}
\end{figure*}

\section{Text-Label Classification} 
In this section, we conduct text-label experiments to prove that MKT can handle text-label inputs better. 
We build a text-label dataset since no such dataset is available.
Specifically, we divide Open Images~\cite{openiamge} dataset into 3830 single-word labels and 3756 text (\ie, multi-words) labels. 
We compare our MKT with BiAM, which is considered as SOTA in ML-ZSL.  
The overall results are shown in Table~\ref{tab:text}, from which we see that our MKT significantly outperforms BiAM in terms of F1 and mAP. 
This is mainly because our MKT can handle a text label as a whole rather than treat it word by word separately like BiAM does.

\section{Comparison of Label Embedding}
\subsection{Label Selection}
In order to compare language model based label embeddings with VLP model based embeddings, we conduct both numerical and visual experiments on part of common labels in NUS-WIDE. 
As presented in Table~\ref{tab:labellist}, we first select 62 typical labels from both 925 seen labels and 81 unseen labels, then divide them into 14 major categories based on their semantic and visual similarity.
For instance, labels ``\textit{kid}'', ``\textit{girls}'', ``\textit{mother}'', ``\textit{man}'', ``\textit{tourist}'' and ``\textit{person}'' all refer to human beings and have similar visual characteristics. 
Due to the visual similarity among these labels, their image embeddings generated by the same vision encoder tend to be similar.
To facilitate the classification based on the similarity between image and label embeddings, their label embeddings are expected to be similar as well.
In Figure 3 of the main paper, the results of the label retrieval experiment demonstrate that VLP model-based label embedding is capable of capturing both semantic and visual consistency between labels. 

\subsection{t-SNE Visualization}
To further illustrate the difference between language model based and VLP model based label embeddings, we utilize t-SNE to visualize the label embeddings generated by GloVe~\cite{glove}, Bert~\cite{devlin2018bert} and prompt-tuned VLP model.
In Figure~\ref{fig:labelemb}, we use different colors and shapes to indicate major categories and types of labels.
As mentioned above, human labels, such as ``\textit{man}'', ``\textit{kid}'' and ``\textit{mother}'', have visual similarities, but language models cannot recognize such relationship among them, causing these labels to be mapped far apart in the embedding space.
The VLP model, on the other hand, is capable of capturing both semantic and visual consistency between these similar labels, resulting in a better distribution of the embedding space.

\begin{table}[t]
\footnotesize
\caption{
Impact of incorrect predictions on AP. 
Last row shows the average precision (AP) of different labels. \underline{Underline} denotes the ground truth label.  \textbf{Bold} denotes the difference of predicted values between label ``White'' and ``Black'', and between instance ``A'' and ``C''. }
\centering
\begin{tabular}{c|cccc}
\toprule[0.15em]
\multirow{2}{*}{\textbf{Sample}} & \multicolumn{4}{c}{\textbf{Prediction of Label}} \\ 
& Dog & Cat & White & Black \\ 
\toprule[0.15em]
A & \underline{0.8} & 0.4 & \underline{0.6} & \textbf{0.7} \\ 
B & \underline{0.3} & 0.6 & \textbf{0.5} & \underline{0.2} \\ 
C & \textbf{0.5} & \underline{0.8} & \underline{0.4} & \underline{0.6} \\
D & 0.6 & \underline{0.1} & \underline{0.2} & \underline{0.4} \\ 
\hline
\textbf{AP} & 0.75 & 0.75 & 0.806 & 0.639 \\ 
\toprule[0.15em]
\end{tabular}
\label{tab:wrong}
\end{table}
\section{Evaluation Metrics}
\subsection{Mean Average Precision}
Following previous work~\cite{Veit_2017_CVPR}, to compute mAP score, we calculate average precision for each class $c$ as
\begin{equation}
A P_{c}=\frac{\sum_{n=1}^{N} \operatorname{Precision}(n, c) \cdot \operatorname{rel}(n, c)}{N_{c}},
\end{equation}
where $\operatorname{Precision}(n, c)$ is the precision for class $c$ when retrieving $n$ highest-ranked predicted scores and $\operatorname{rel}(n, c)$ is an indicator function that is 1 iff the image at rank $n$ contains label $c$ and 0 otherwise.
$N_{c}$ denotes the number of positives for class $c$.
Then mAP is computed as 
\begin{equation}
m A P=\frac{1}{d} \sum_{c=1}^{d} A P_{c},
\end{equation}
where $d$ is the number of labels.
\subsection{F1 score}
Following previous work~\cite{gong2014deep}, we assign $n$ highest-ranked predictions to each image and compare them with the ground truth labels. 
The mean-per-label precision and mean-per-label recall are defined as 
\begin{equation}
P \triangleq \frac{\sum_{c} N_{c}^{t}}{\sum_{c} N_{c}^{p}}, \quad R \triangleq \frac{\sum_{c} N_{c}^{t}}{\sum_{c} N_{c}},
\end{equation}
where $N_{c}^{t}$ is the number of true positive for label $c$ and $N_{c}^{p}$ is the number of positive predictions for label $c$. Then, the F1 score is computed as
\begin{equation}
F 1=\frac{2 P R}{P+R}.
\end{equation}

\subsection{Discussion of Metrics}
As discussed in Section 4.1 of the main paper, mAP reflects the ranking accuracy of each label across all images, and F1 score reflects the label ranking accuracy of each image.
By comparison to F1 scores, we argue that mAP is more sensitive to wrong predictions with high scores.
For instance, in Table~\ref{tab:wrong}, the difference of predicted values between label ``\textit{White}'' and ``\textit{Black}'' are the values of negative labels.
When other values are identical, the higher wrongly predicted value will result in a lower mAP score.
On the other hand, the difference of predicted values between instances ``A'' and ``C'' also corresponds to the negative labels. 
However, despite the higher wrongly predicted value, the Top-3 prediction accuracy and F1 score are not affected.
There are two reasons for this.
First, when calculating the F1 score, the Top-3 predictions are treated equally and their relative order is not taken into account.
However, when computing mAP, the ranking of every predicted value for a label matters.
Meanwhile, as illustrated in Figure 4 of the main paper, the distribution of higher values is more dispersed, and minor perturbations may not change \textit{top}-$K$ predicted labels.
However, the same perturbations on smaller value predictions for negatives may change their rankings, resulting in a lower mAP score. 

\begin{table}[t]
\caption{Comparison of Complexity. For a fair comparison, both models are run on the Tesla V100.}
\centering
\footnotesize
\begin{tabular}{ccccc}
\toprule[0.15em]
\textbf{Method} & \textbf{mAP} & \textbf{Inference} & \textbf{FLOPs} & \textbf{Params} \\
\hline
BiAM & 25.9 & 8.3 ms  & 20.19 G & 143.4 M \\
MKT & 37.6 & 13.8 ms & 16.99 G & 86.8 M \\
\toprule[0.15em]
\label{tab:complex}
\end{tabular}
\end{table}
\begin{figure}[t]
  \centering
  \subcaptionbox{GZSL\label{fig:dis_gzsl}}
    {\includegraphics[width=0.49\linewidth, page=1]{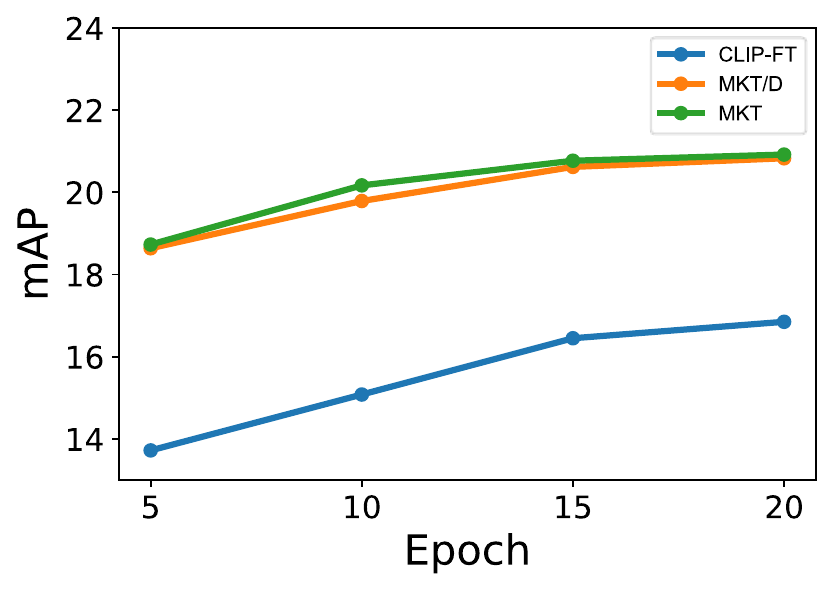}}
  \subcaptionbox{ZSL\label{fig:dis_zsl}}
    {\includegraphics[width=0.49\linewidth, page=2]{images/procedure.pdf}}
  \caption{Comparison of training procedures. Results are reported in terms of mAP on NUS-WIDE dataset. ``MKT/D'' means that we train the MKT without distillation. 
  }
  \label{fig:dis}
\end{figure}

\begin{figure*}[ht]
    \centering
    \includegraphics[width=1\textwidth, page=1]{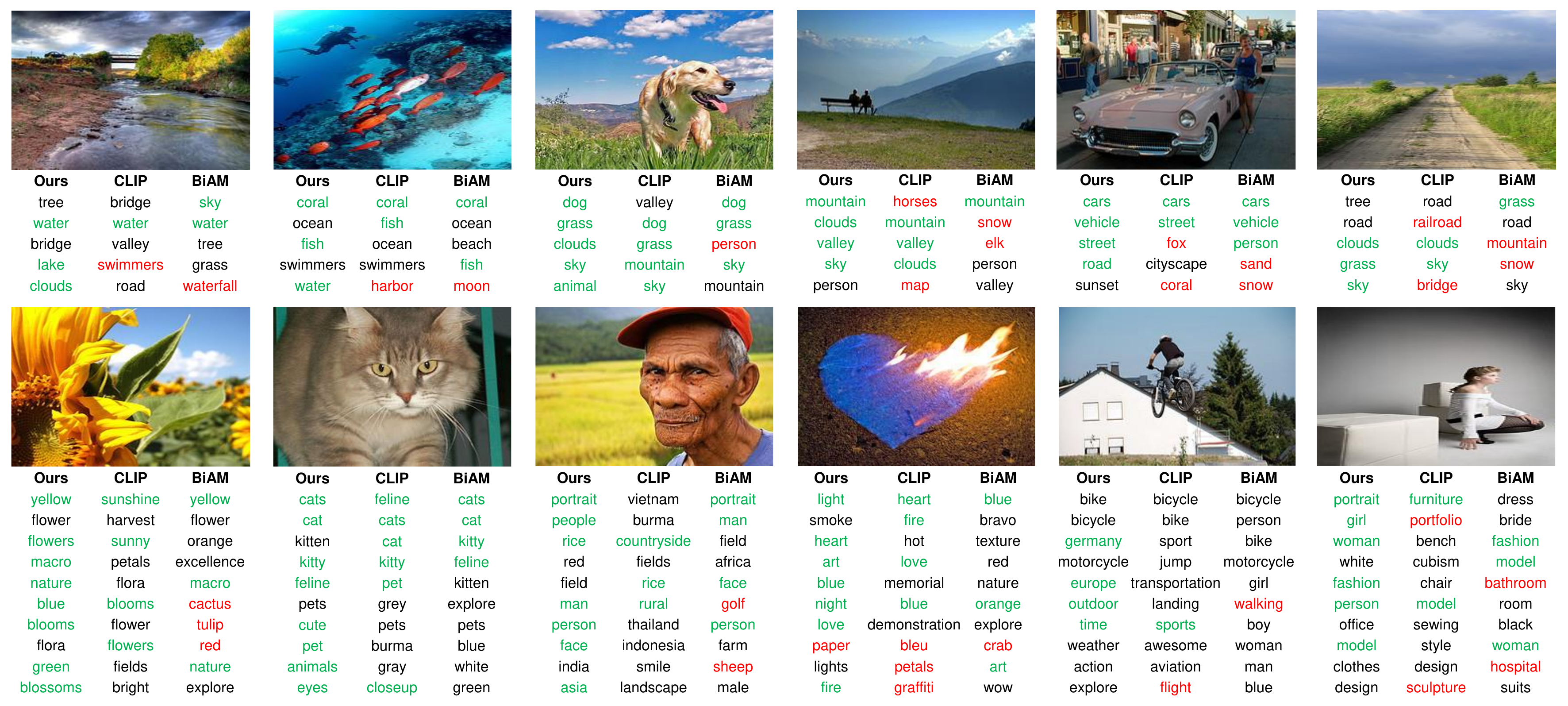}
\caption{
Comparison of predictions among our model, BiAM, and CLIP. 
The top row shows the Top-5 predictions in ZSL task, while the bottom is the Top-10 predictions in GZSL task. 
True positive predictions are shown in green and red font denotes incorrect predictions. 
}
    \label{fig:prediction_supp2}
\end{figure*}

\begin{figure}[ht]
\centering
\includegraphics[width=1\columnwidth, page=3]{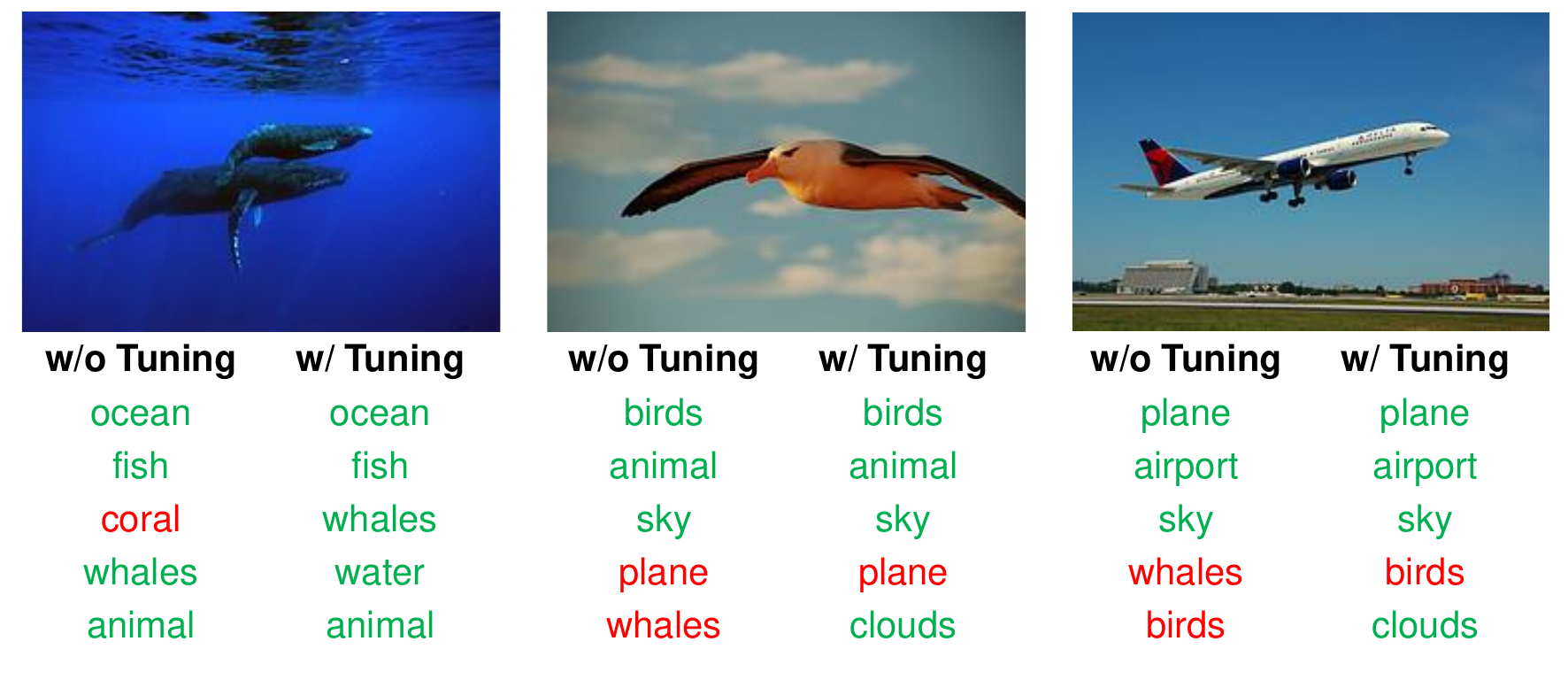}
\caption{
Comparison of predictions between the first and second stage of our method.
True positive predictions are shown in green and the red font denotes apparently incorrect predictions. }
\label{fig:prediction_supp}
\end{figure}


\section{Comparison of Complexity}
\label{section:Comparison of Complexity}
We compare the complexity of our model and BiAM in Table~\ref{tab:complex}.
In BiAM~\cite{biam}, they did not take the complexity of backbones into account.
As shown in Table~\ref{tab:complex}, compared with BiAM, MKT achieves superior performance with fewer parameters and FLOPs.

\section{Effect of Knowledge Distillation}
To further explore the effect of knowledge distillation, we compare the performance during the training procedure between MKT and CLIP-FT.
CLIP-FT is a pre-trained CLIP model and fine-tuned on seen labels with ranking loss.
For a fair comparison, in this experiment, we initialize the vision backbone in MKT with pre-trained CLIP~\cite{clip}. 
Results are presented in Figure~\ref{fig:dis}. 
During the training procedure, without distillation (\eg, CLIP-FT and MKT/D), performance improves consistently on GZSL task but declines on ZSL task, indicating severe overfitting.
As a result of knowledge distillation, MKT achieves superior performance on both ZSL and GZSL tasks.
We argue that knowledge distillation can alleviate the overfitting problem and facilitate the alignment between image and text embeddings.
Additionally, we find that even without distillation, the MKT/D model outperforms the fine-tuned CLIP model on both ZSL and GZSL tasks, demonstrating MKT's effectiveness for multi-label classification.

\section{Effect of Prompt Tuning}
Since handcrafted prompts cannot be guaranteed to be optimal for our task, we use prompt tuning to adapt the context embedding further.
We argue that through prompt tuning, label embedding tends to pay more attention to visual information.
For instance, as illustrated in Figure 3b of the main paper, the Top-3 retrieval results of query ``\textit{girls}'' on CLIP embedding are ``\textit{man}'', ``\textit{kid}'' and ``\textit{school}''.
Label ``\textit{school}'' and ``\textit{girls}'' have semantic relevance, but few visual similarity.
In contrast, the Top-3 retrieval results of query ``\textit{girls}'' on Prompt embedding are ``\textit{man}'', ``\textit{kid}'' and ``\textit{person}''.
As these labels have a similar visual appearance, their embeddings should also be similar.
Figure~\ref{fig:prediction_supp} shows the Top-5 predictions in the first and second stage of our method on ZSL task.
Notice that the images containing label ``\textit{whales}'' usually consist of a blue background and a huge foreground object.
The first-stage model tends to confuse whales with a bird or plane in the sky.
Hence, prompt tuning may also facilitate the classification of similar categories by making the embedding of visually similar labels more distinguishable. 

\section{Additional Qualitative Results}
\label{section:Additional Qualitative Results}
To further demonstrate the superiority of our model, we present an additional comparison of predicted results on NUS-WIDE among CLIP, BiAM and our approach.
Figure~\ref{fig:prediction_supp2} shows the Top-5 and Top-10 predictions in ZSL and GZSL tasks respectively.
MKT is capable of recognizing visual and semantic consistency as well as capturing local and global features, resulting in more precise and diverse predictions.

\end{document}